\documentclass[letterpaper, 10 pt, conference]{ieeeconf}  

\IEEEoverridecommandlockouts                              

\usepackage{graphics} 
\usepackage{epsfig} 
\usepackage{mathptmx} 
\usepackage{times} 
\usepackage{cite}
\usepackage{amsmath,amssymb,amsfonts}
\usepackage{algpseudocode}
\usepackage{algorithm}
\usepackage{graphicx}
\usepackage{textcomp}
\usepackage{xcolor}
\usepackage{censor}
\usepackage{graphicx}
\usepackage{subcaption}
\usepackage{multirow}
\usepackage[export]{adjustbox}
\usepackage{commands}

\title{\LARGE \bf
Shared-Roadmap Generation and Evaluator for Multi-Agent Path Planning Using Heterogeneous Graph Neural Network
}

\author{Brandon Ho, Nikola Rogers, Seung-Kyum Choi
\thanks{Brandon Ho, Nikola Rogers, and Seung-Kyum Choi are with the Institute of Robotics and Intelligent Machines (IRIM), Georgia Institute of Technology, Atlanta, GA, USA 
        {\tt\small bho36@gatech.edu}}%
}

\begin{document}
\maketitle
\begin{abstract}
Multi-agent path planning (MAPP) in continuous environments often relies on roadmaps to balance safety and search efficiency. 
However, traditional roadmap generation methods, such as lattice grids or standard sampling-based approaches, frequently face a trade-off between graph density and the likelihood of finding feasible, high-quality solutions. 
In this paper, we propose a scalable heterogeneous Graph Neural Network (GNN) framework for the automated generation and evaluation of shared multi-agent roadmaps. 
Our model covers the representation of waypoints, agent locations, and task locations as distinct nodes in a heterogeneous graph, allowing it to reason over global connectivity and inter-agent interactions. 
By training on occupation density maps aggregated and collected from expert solver trajectories, the GNN learns to identify critical points of interest and prune redundant nodes and edges. 
This process produces a compact, coordination-aware roadmap that is invariant to task permutations and is reusable for multi-agent pick and delivery tasks. 
Experimental results demonstrate that our framework can reduce planning effort and can potentially find better solutions, reaching at least $40\%$ reduction in runtime and in graph size for dense roadmaps.
\end{abstract}
\section{Introduction}


Multi-agent path planning (MAPP) has been a central challenge in domains such as autonomous driving \cite{gonzalez_review_2016, khayatian_survey_2020}, aerial swarms \cite{kim_increasing_2021}, and warehouse automation \cite{ho_towards_2025, andrade_lets_2025}, where overall system performance depends critically on both safety and efficiency.
In real-world deployments, safety requirements typically take precedence over efficiency, often constraining agents to move along predefined edges and vertices within a structured environment, commonly referred to as a shared roadmap.
A simple solution is a lattice grid, where autonomous agents or automated guided vehicles (AGVs) are restricted to moving between cardinal neighboring cells.
However, unlike lattice grids, roadmaps that approximate continuous state spaces allow agents to move more flexibly and can yield more practical and higher-quality solutions. 
Such non-lattice roadmaps better capture the underlying geometry of real environments and reduce discretization artifacts. Nevertheless, roadmap generation itself is a non-trivial problem. 
Poorly constructed roadmaps can significantly degrade the performance of MAPP, underscoring the importance of principled methods for a shared multi-agent roadmap generation.

The main challenge of roadmap generation lies in balancing graph density with the likelihood of finding a feasible solution \cite{alkazzi_comprehensive_2024}.
A dense roadmap may better capture high-quality or even the optimal solution, but it enlarges the search space.
Conversely, a sparse roadmap might expedite the search process but risk of finding a valid solution.

A common approach for constructing effective roadmaps is a two-step iterative process: (1)  use sampling-based motion planning (SBMP) \cite{elbanhawi_sampling-based_2014} to generate multiple roadmaps and then (2) evaluate these roadmaps using a MAPP algorithm.
Although this paradigm has been widely adopted for single-agent navigation \cite{kumar_lego_2019,jia_planar_2014 ,zhou_grid-based_2024, yu_reducing_2022}, extending it to multi-agent settings is non-trivial due to the inter-agent interactions.
While prior work has explored roadmap generation for multi-agent scenarios \cite{henkel_optimized_2020, okumura_ctrms_2022, arias_avoidance_2021}, existing approaches impose certain constraints, such as generating non-shared roadmaps for each agent, limiting connectivity (e.g., non-bidirectional edges).

\begin{figure*}
    \centering
    \includegraphics[width=\linewidth]{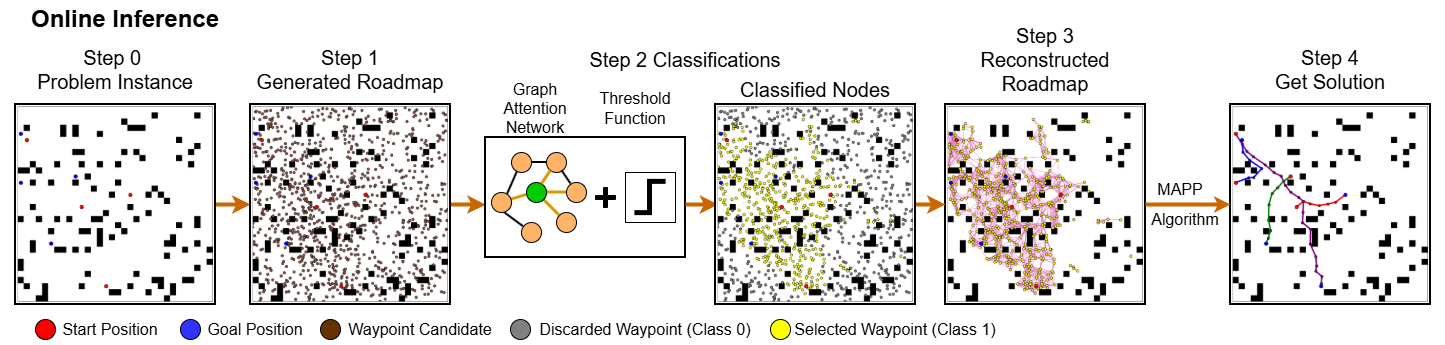}
    \caption{Overview of the proposed learning-based roadmap evaluator and reconstruction framework for multi-agent path planning (MAPP).  Given a new problem instance, a dense roadmap is generated from any roadmap generator and is then evaluated by the trained GNN. A thresholding function classifies vertices as selected or discarded waypoints. The selected nodes form a reconstructed roadmap, which is subsequently used by a MAPF/MAPP algorithm to compute the final solution.}
    \label{fig:method}
\end{figure*}

Graph Neural Networks (GNNs) have shown promising results in scalable recommendation systems where information can be represented in a relational graph \cite{wu_graph_2022}.
By coding individual entities as nodes and their interactions as edges, GNNs are able to learn structured representations that generalize to environments with a variable number of elements. 
Given an existing roadmap, we encode waypoints, agent locations, and task locations as nodes in a heterogeneous graph, while encoding spatial relationships and task-relevant information as edge features. 
On this structured representation, our heterogeneous GNN model evaluates roadmap quality and reasons over global connectivity and agent interactions. 
Additionally, the GNN learns to identify critical points of interest and prune less informative nodes or edges, thereby producing a refined roadmap that is both compact and better suited for efficient multi-agent coordination.
Figure \ref{fig:method} showcases our proposed framework.

Our contributions are as follows:
\begin{itemize}
    \item We propose a dynamic and scalable heterogeneous Graph Neural Network (GNN) framework for roadmap generation and evaluation. Our model identifies critical points of interest within an existing roadmap and prunes less informative nodes and edges, producing a more compact and coordination-aware graph.
    \item We design our model to be task-agnostic for multi-agent path planning (MAPP) and readily extensible to multi-agent pickup and delivery (MAPD). By representing agent locations and task locations under the same node type as landmarks, the constructed roadmap becomes invariant to task configuration permutations, enabling flexible deployment across varying task assignments
    \item We demonstrate the potential of our approach by training on MAPP instances up to four agents and testing up to 16 agents. Experimental results show that our GNN-based framework reduces planning effort, achieving at least a $40\%$ reduction in runtime and graph size for dense roadmaps, and can potentially find better solutions.
\end{itemize}


\section{Related Works}

\subsection{Multi-Agent Pathfinding (MAPF)}
Multi-agent pathfinding (MAPF), a problem of multi-agent path planning (MAPP) in discrete space, has been applied in various AI and robotics applications where the objective is to find a set of collision-free paths on a graph.
Popular approaches include, but are not limited to, conflict-based search (CBS) algorithms \cite{sharon_conflict-based_2015}, increasing cost tree search \cite{walker_extended_2018}, and prioritized planning (e.g., HCA* \cite{silver_cooperative_2005}, SIPP \cite{phillips_sipp_2011}, etc.
More recently, research demonstrates the potential in using machine-learning techniques for solving MAPF.
These methods learns from either demonstrations collected offline for imitation learning, a sequence of observations for reinforcement learning, or using graph neural networks, leading to finding solutions in multi-agent path planning (MAPP).
Despite such progress, many algorithms simplify the scope of their problem instances, typically in a grid map, where the spatial graph structure has uniform travel cost, and collisions only occur when point-particle agents occupy the same vertices or edges \cite{silver_cooperative_2005}.
These assumptions restrict the algorithms' capabilities to only discrete time and space, making their real-world applications limited.

\subsection{Roadmap Generation}
Roadmap generation has been a research topic of interest mainly for single-agent cases, as it allows the agents to move in continuous space, leading to more practical and better solutions.
A roadmap refers to an undirected connected graph that is either shared across agents or tailored for each agent.
An ideal roadmap entails that any two points on the graph can be connected by at least one path, and the graph should be sparse as possible.
Popular methods in generating a roadmap include, but not are not limited to, probabilistic roadmaps (PRMs) \cite{kavraki_probabilistic_1996}, constrained Delaunay Triangulation (CDT) \cite{kallmann_fully_2004}, visibility graph \cite{huang_dynamic_2004}, and rapidly exploring random graphs (RRGs).
CDT and visibility graph construct a roadmap by utilizing the obstacle boundary as either constraints or points of interest, while PRM and RRG iteratively sample points in free space.
Research on single-agent roadmaps studies how to densify the representation of the environment to balance optimality and density \cite{kumar_lego_2019,jia_planar_2014 ,zhou_grid-based_2024, yu_reducing_2022}.

For multi-agent roadmaps, most continuous MAPP methods \cite{
andreychuk_improving_2021,
wahba_pc-dbcbs_2025,ali_safe_2023,
kasaura_prioritized_2022,
moldagalieva_db-lacam_2025
} rely on some roadmap to operate.
Their primary challenge lies in scaling to large numbers of agents in real time, as the search space grows significantly and collision checking must be performed at high frequency.
In contrast, we investigate a complementary research direction: instead of improving the planner itself, we explore what types of roadmap representations are most effective for solving MAPP in continuous space.

There have been relatively few studies on multi-agent roadmap generation and refinement.
\cite{arias_avoidance_2021} constructs a shared roadmap with bias-based sampling, focusing on narrow passages, but is less suited for sparsely distributed obstacles.
\cite{okumura_ctrms_2022} employs a generative model to construct separate but not shared roadmaps per agent that incorporate inter-agent interactions.
The most relevant work is \cite{henkel_optimized_2020}, where they optimized already constructed roadmaps through online learning.
However, their optimization is limited to directed edges and does not remove redundant or unnecessary vertices, thereby restricting the extent to which the roadmap can be structurally refined.

In contrast, our method identifies critical points of interest within an existing roadmap and reconstructs a shared, refined graph.
A shared roadmap provides constraints where safety can be better tracked as the agents follow on set guided paths.
By labeling agents’ initial and task locations as landmarks, the model ensures connectivity among essential nodes, improving coverage, reusability, and generalization across varying multi-agent tasks, including pickup and delivery scenarios. 
\section{Problem Statement} \label{sec:problem_statement}
\textbf{Problem Description}: A  MAPP  problem instance $\probleminstance$ is defined as a tuple $(\goalset, \startset, \agentspace, \classspace, \radiusspace, \graph, \kinematicspace)$.
$\graph = (\vertices,\edges)$ is an undirected roadmap graph constructed in the state space $\classspace$ with vertices $\vertex \in \vertices$, representing position agents can occupy, and edge $\edge_{ij} \in \edges: (\vertex_i,\vertex_j)$ indicating traversal connections between $\vertex_i$ and $\vertex_j$ without collision, i.e., $\edge_{ij} \in \classspace_{\obs}$.
We consider a MAPP for a team of $\numberof$ agents $\agentspace = \{\agent, ..., \agent_\numberof\}$ in a $2D$ continuous space $\classspace \subseteq \realnumber^2$, where $\classspace_{\obs} \subseteq \classspace$ represents the obstacle set, and the remaining part is the free space $\classspace_{free} \subset \classspace \backslash  \classspace_{obs}$.

Each agent $\agent_i$ has a unique start position $\start_i$ and goal position $\goal_i$, where $\startset: \{\start_i\}_{i\in\{1,...\numberof\}}$ and $\goalset: \{\goal_i\}_{i\in\{1,...\numberof\}}$ represent the start vertices set and goal vertices set respectively.
Each agent's body is modeled as a ball with radius $\radius_i(q) \subseteq \classspace$ around its position $q \in \classspace$ where $\radiusspace: \{\radius_i\}_{i=\{1,...,\numberof\}}$ represents the space occupied by the agent .
Each agent $\agent_i$ has its kinematic constraints $\kinematic$, e.g each agent has a maximum velocity or move distance where $\kinematicspace = \{\kinematic\}_{i=\{1,...,\numberof\}}$.
A trajectory for agent $\agent_i$ is defined by a continuous mapping $\sigma_i: [0,T_i] \rightarrow \classspace$ where $T_i \in \realnumber_{>0}$.

Note that the agent will stay in place after reaching the goal point, starting from $t \geq T_i$, which means that $\sigma_i(t) = \sigma_i(T) \forall t \geq T_i$. A solution for a MAPP problem instance is a set of $\numberof$ trajectories $\trajectories = \{\trajectory_i, ..., \trajectory_\numberof\}$ that satisfies the following conditions:
\begin{enumerate}
    \item  \textbf{Endpoint}:  $\trajectory_{i}(0) = \start_i \wedge  \trajectory_{i}(T) = \goal_i $ 
    \item \textbf{Obstacles}:   $\trajectory_{i}(t) \in \classspace_{free}, 0 \leq t \leq T $ 
    \item \textbf{Inter-agent}: $\radius_i(\trajectory_{i}(t)) \cap \radius_i(\trajectory_{j}(t)) = \emptyset, 0 \leq t \leq T $ 
    \item \textbf{Kinematic-aware}: $\sigma_i$ satisfies constraint $\kinematic_i$
\end{enumerate}

\textbf{Solution Cost}: We defined the quality of the solution $\trajectories$ by \textit{sum-of-costs:} $\sum_iT_i$ for $i \in \{1,...,\numberof\}$ which represents the sum of all time taken for each agent $\agent_i$ to reach their goal.

\section{Methodology}
In this section, we first describe the workflow of the heterogeneous graph neural network model, starting with the roadmap generation, then the model architecture, and its pruning and reconstruction process.

\subsection{Roadmap Generation}
\begin{figure}[ht]
    \centering
    \begin{subfigure}{.225\textwidth}
        \centering
        \includegraphics[width=\linewidth]{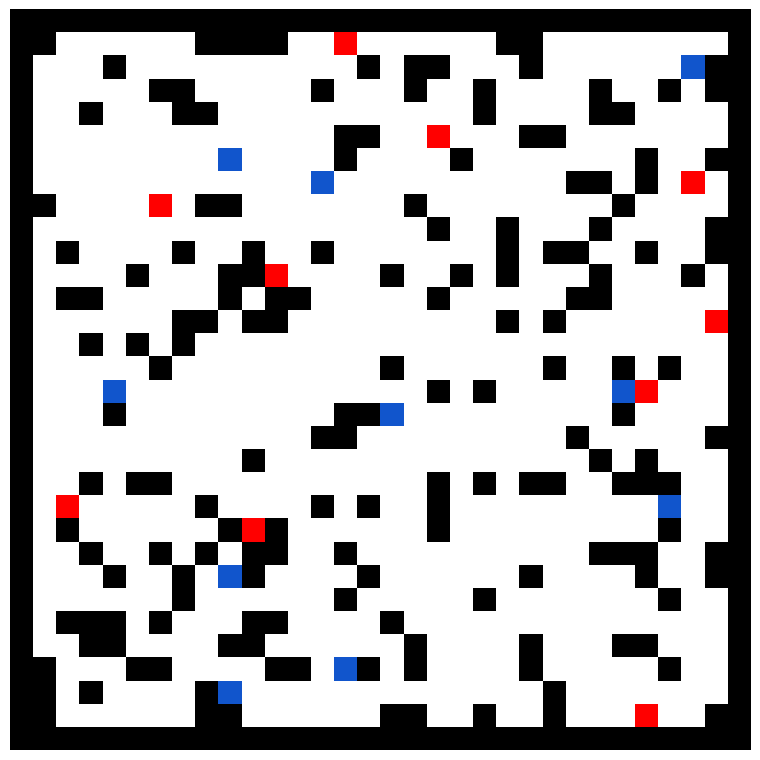}
        \caption{Map}
        \label{subfig:roadmaps_map}
    \end{subfigure}%
    \hspace{0.02\textwidth}
    \begin{subfigure}{.225\textwidth}
        \centering
        \includegraphics[width=\linewidth]{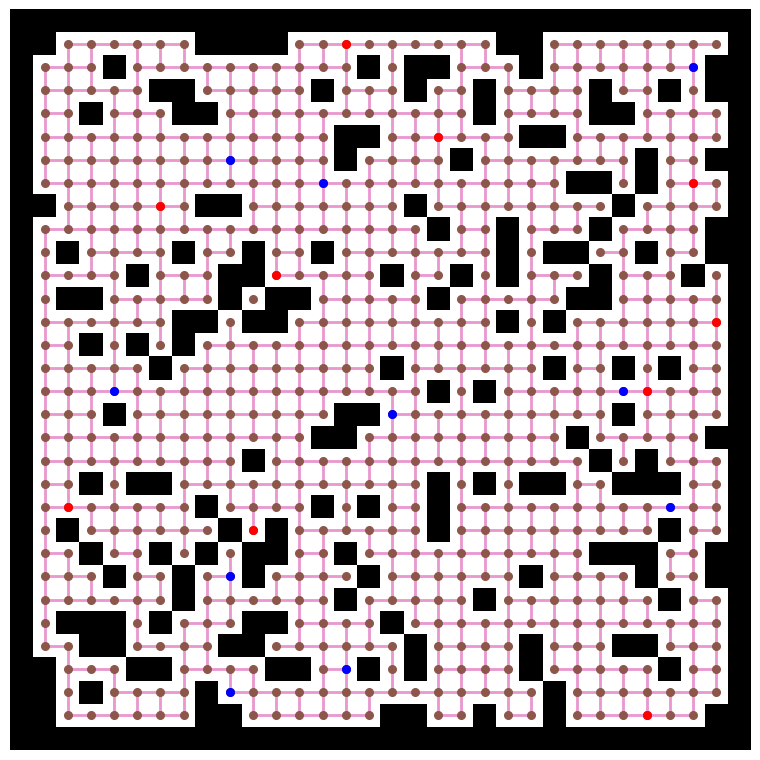}
        \caption{Lattice Grid}
        \label{subfig:roadmaps_lattice}
    \end{subfigure}

    \begin{subfigure}{.225\textwidth}
        \centering
        \includegraphics[width=\linewidth]{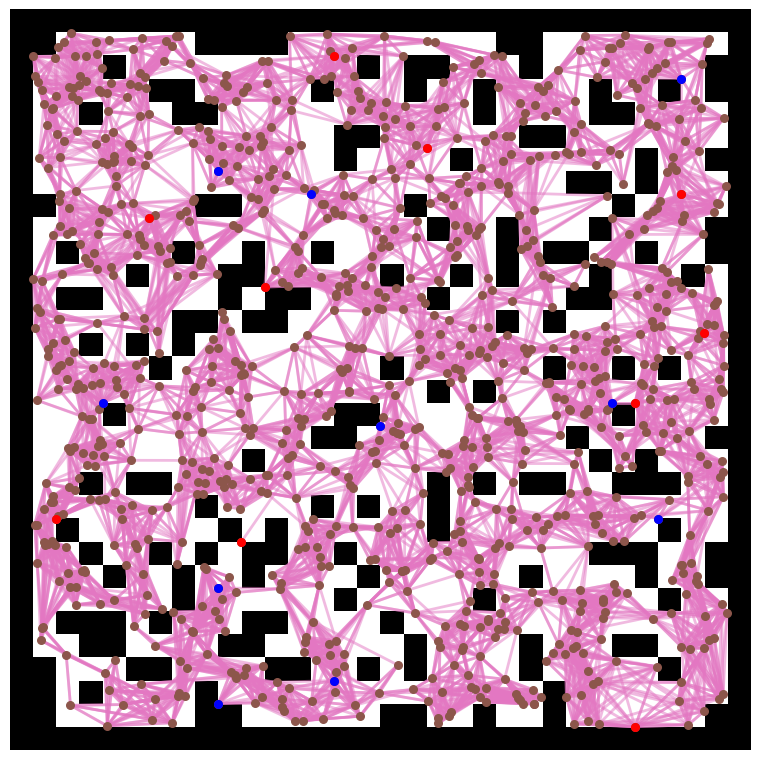}
        \caption{PRM}
        \label{subfig:roadmaps_prm}
    \end{subfigure}%
    \hspace{0.02\textwidth}
    \begin{subfigure}{.225\textwidth}
        \centering
        \includegraphics[width=\linewidth]{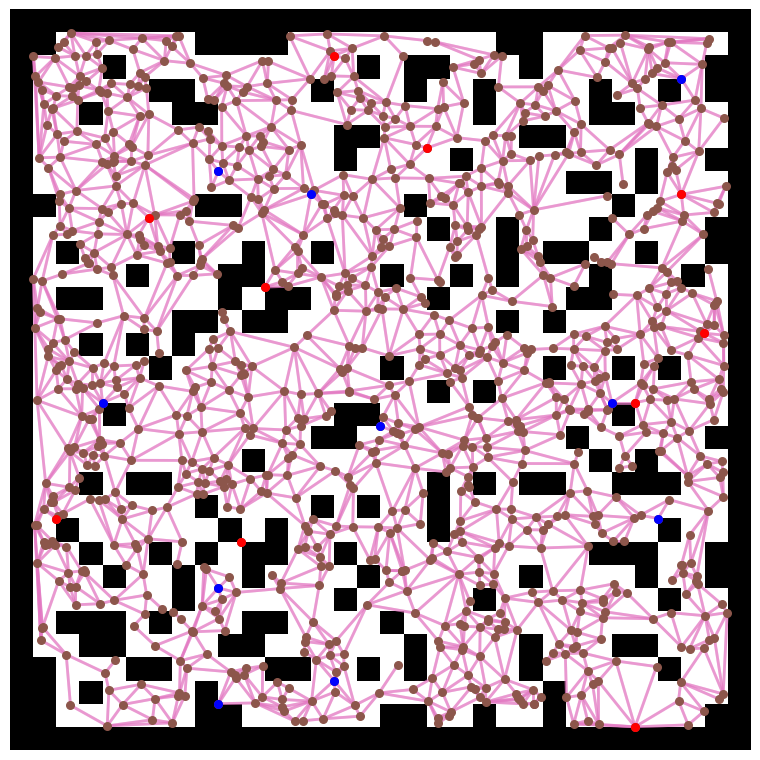}
        \caption{Delaunay Triangulation}
        \label{subfig:roadmaps_dt}
    \end{subfigure}
    \caption{Comparison of the original map (Fig. \ref{subfig:roadmaps_map}) and the roadmaps (\ref{subfig:roadmaps_lattice}-\ref{subfig:roadmaps_dt}). The red and blue represent the start and goal points, respectively.}
    \label{fig:roadmaps}
\end{figure}

We construct our homogeneous roadmap graph $\graph$ using three generation strategies, as illustrated in Fig.~\ref{fig:roadmaps}.
The first strategy is a lattice grid, in which vertices are uniformly distributed throughout the free space and edges connect neighboring vertices along Cartesian directions.
The second strategy is a Probabilistic Roadmap (PRM), where vertices are randomly sampled in the free space and edges are created based on a maximum connection radius and a limit on the number of nearest neighbors.
The third strategy is Delaunay triangulation, in which randomly sampled vertices are connected to form a planar graph composed of non-overlapping triangles.
For all three constructions, each vertex $\vertex_i \in \graph$ stores its global position a, and each edge $\edge_{ij} \in \graph$ is associated with a scalar weight $\weight_{ij}$, corresponding to the Euclidean distance between its incident vertices. 

\begin{figure*}
    \centering
    \includegraphics[width=\linewidth]{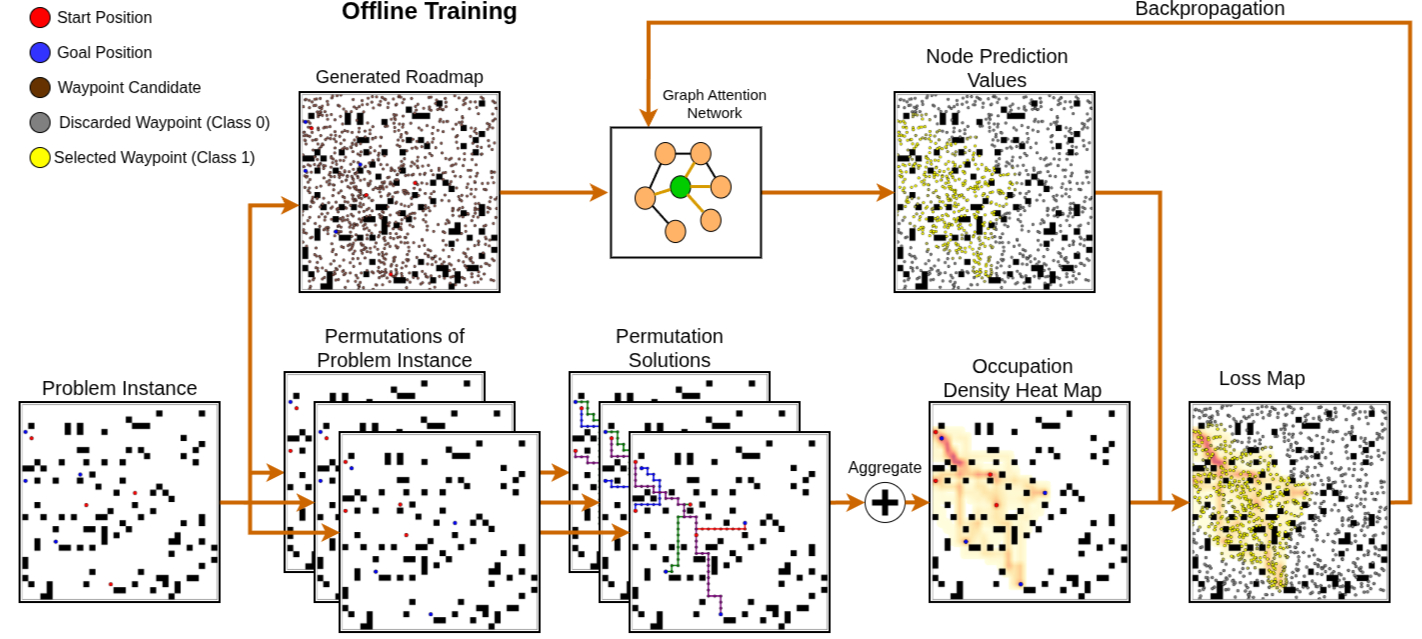}
    \caption{Offline training process. For each new problem instance in the dataset, randomized start-goal assignments are generated to create multiple permutations of the same problem instances. These permutations are then solved to generate solution trajectories, which are then aggregated into an occupation-density heatmap. This density map provides supervision to train a graph neural network (GNN) to predict node importance value.}
    \label{fig:placeholder}
\end{figure*}

\subsection{Model}
In this work, we formulate roadmap evaluation and generation as a binary node classification problem e.g ($\phi(\vertex_i | \graph) \rightarrow \{0,1\}$).
We are given a collection of MAPP demonstrations, each consisting of a problem instance and its corresponding solution obtained through computationally intensive offline MAPF solvers operating on sufficiently dense roadmaps constructed via uniform random or grid sampling.
Using these demonstrations as training data, we learn a predictive model that classifies each vertex according to whether it belongs to the solution set (class 1) or not (class 0). 
The trained model serves as a vertex evaluator that prunes a densely generated shared roadmap based on a decision threshold, producing a sparser yet solution-preserving graph that enables more efficient downstream planning.
Figure \ref{fig:method} showcased our proposed framework.

\subsubsection{Architecture} \label{subsubsec:architecture}
\hfill

\textbf{Graph Augmentation}: The GNN evaluator $\phi$ operates on an augmented heterogeneous graph $\mathcal{G}_{\text{aug}}$ derived from the initial roadmap $\mathcal{G}$.
To make the network task-aware, node features in $\mathcal{G}$ are concatenated with two-bit binary encodings indicating membership in the start set $\mathcal{S}$ or goal set $\mathcal{G}$. 
These encodings guide the GNN to identify critical landmarks. Furthermore, we introduce a global directed edge type for $\graph_{\text{aug}}$ that connects every landmark $\vertex_i \in \mathcal{S} \cup \mathcal{G}$ to all other vertices $\vertex_j \in \mathcal{V}$. 
Each new edge $\edge_{ij} = (\vertex_i, \vertex_j)$ is weighted by the Euclidean distance between the vertices. 
This secondary connectivity layer provides the GNN with global contextual information, aiding in the identification of waypoints likely to belong to the optimal solution set.

\textbf{Model Layers}: Before the GNNs, the node and edge features are first encoded into an embedding with two linear layers. 
Afterward, the node and edge embedding are passed into three residual multi-head Graph Attention Networks (GAT) operators \cite{velickovic_graph_2018, zhang_gresnet_2019}, followed by a leaky ReLU activation function.
The vertex representation is then processed through another linear layer, followed by a sigmoid function.
The output $\outputvalue =\phi(\graph_{\text{aug}})$ predicts a scalar value per vertex, representing the probability of how likely that vertex is used by the solution set.

\subsubsection{Training and Data Generation}
\hfill

\textbf{Ground Truth}:
Given a MAPF instance, we first apply Conflict-Based Search (CBS) to compute a feasible solution on a lattice-grid roadmap. 
Regardless of whether CBS successfully finds a solution, we generate additional training variations by randomizing the start and goal assignments within the same environment.
For each modified MAPF instance, we again apply CBS to obtain a feasible solution when possible.
This process is repeated until a sufficient number of feasible start–goal permutations are collected.
After obtaining multiple solution trajectories, we aggregate the agents’ paths spatially to construct an occupation density map, which captures the frequency with which different regions of the roadmap are utilized across demonstrations. 
This occupation density map $H$ serves as ground truth and a labeler for the data generation

\textbf{Data Generation}:
Given the same MAPF instance, we construct a roadmap $\graph$ using PRM.
Each vertex $v_i \in \graph$ is labeled according to its relevance to the occupation density $H$.
If a vertex lies directly on the heatmap (i.e., it was visited in the aggregated demonstrations), it is assigned a label of 1. 
Otherwise, its label is determined based on proximity to labeled vertices through a nearest-neighbor propagation scheme.
The labeler scoring function $f$ is defined as:
\begin{equation}
    f(\vertex_i) = \begin{cases}
              1 & \vertex_i \in \heatmap \\
              \max_{j \in \neighborhood(i)}{\frac{f(\vertex_j)}{1+dist(\vertex_i,\vertex_j)}} & \vertex_i \notin \heatmap
            \end{cases}
\end{equation}
where $\neighborhood(i)$ denotes the neighborhood of $\vertex_i$, and $dist(\vertex,\vertex')$ is in the Euclidean distance between vertices.
This formulation assigns a high score to vertices frequently visited in the solutions while smoothing decaying the score for nearby vertices, creating a soft relevance signal over the roadmap.

\begin{figure}[ht]
    \centering
    \begin{subfigure}{.24\textwidth}
        \centering
        \includegraphics[width=\linewidth]{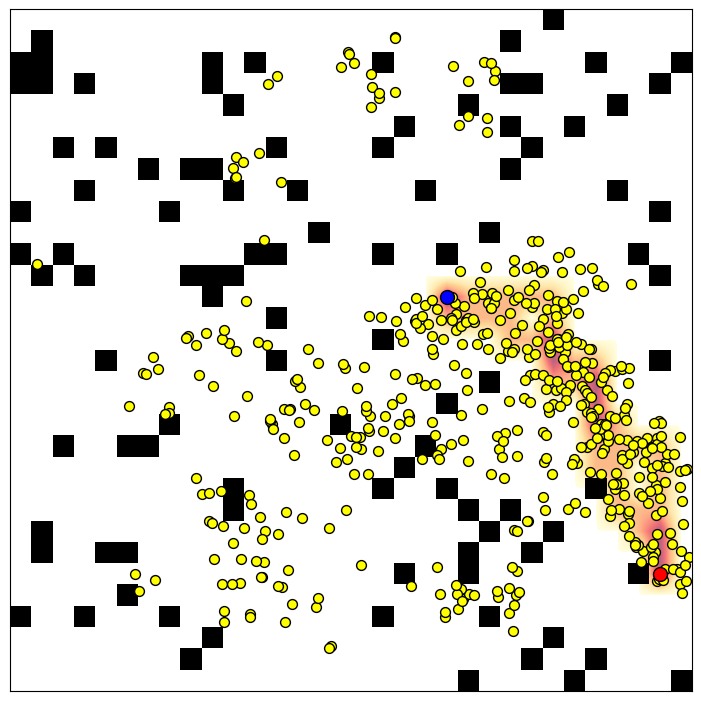}
        \caption{Single Agent Validation}
        \label{subfig:single_agent_valid}
    \end{subfigure}%
    \begin{subfigure}{.24\textwidth}
        \centering
        \includegraphics[width=\linewidth]{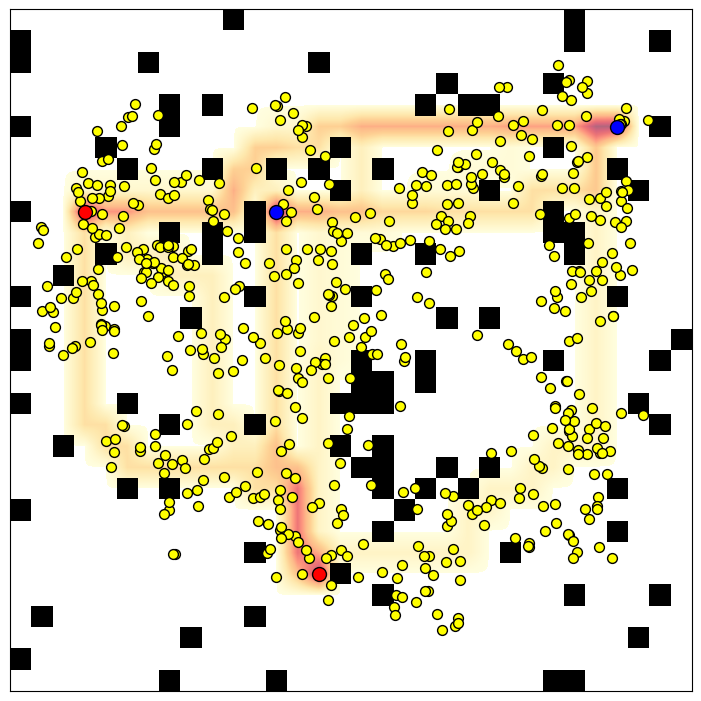}
        \caption{Multi-Agent Validation}
        \label{subfig:multi_agent_valid}
    \end{subfigure}
    \caption{These 32 x 32 validation scenarios show the GNN evaluator's ability in selecting which nodes to be used for reconstruction in (a) single agent scenarios and (b) multi-agent scenarios. The underlying heatmap highlights the agent paths found by CBS for multiple permutations of the same MAPF problem instances. The red and blue are the start and goal locations for this specific MAPF problem instance permutation.}
    \label{fig:validation}
\end{figure}

\textbf{Training}
The objective of the model is to classify how likely the vertex belongs to the solution set. 
We learn the following loss function, which is a combination of the focal loss \cite{lin_focal_2018} and Laplacian loss \cite{li_laplacian-steered_2017}:
\begin{equation}
    \hspace{-5pt}\loss(\graph,\edges,\nnweight) = -\frac{1}{\numberof}\sum_{i=1}^{\numberof} \alpha(1-\outputvalue_i)^{\gamma}\log(\outputvalue_i) +  \frac{\lambda}{\numberof}\sum_{i=1}^{\numberof}\sum^{j\in\neighborhood(i)}_{j=1}{\hspace{-5pt}\weight_{ij}(\outputvalue_i - \outputvalue_j)^2}
\end{equation}
where $\gamma$ and $\alpha$ are focal loss parameters, $\lambda$ is a regularization term, $p_i$ are the output probabilities, $\nnweight$ is the GNN parameters.
Since the dataset is imbalanced, as most vertices do not lie on the solution heatmap for a few agents, we employ focal loss to mitigate the dominance of easy negative samples and emphasize informative vertices.
The Laplacian term encourages smoothness over the graph by penalizing large prediction differences between neighboring vertices, aligning with the spatial decay structure introduced during data generation. 
Figure \ref{fig:validation} shows an example of our GNN evaluator in our validation scenarios.

\subsection{Reconstruction}
In this subsection, we explain how to reconstruct a new roadmap using the GNN as a vertex evaluator.
Algorithm~\ref{alg:reconstruct} summarizes the proposed procedure, which takes a problem instance $\probleminstance$ as input and outputs a newly refined map $\graph_{new}$.
Figure \ref{fig:method} visualizes the process.

First, we generate a dense shared roadmap $\graph$ using a roadmap construction function conditioned on the problem instance $\probleminstance$.
The specific generation strategy is flexible (e.g. PRM), but it must provide a sufficiently dense representation of the environment to ensure coverage.
Next, the roadmap $\graph$ is augmented into a heterogeneous graph representation $\graph_{\text{aug}}$ suitable for input to the GNN $\phi$, as previously discussed in Section \ref{subsubsec:architecture}.
A forward pass through $\phi$ produces vertex probability scores indicating the likelihood that each vertex belongs to the solution set.
These probabilities are then passed to a threshold function to obtain a binary class assignment.
The threshold value ($\tau$) is defined as
\begin{equation}
  \tau(\graph,\textbf{y}) = \min(\text{mean}(\textbf{y}),,0.5),  
\end{equation}
where $\textbf{y}$ denotes the set of predicted vertex scores. 
The reconstructed roadmap $\graph_{new}$ is then obtained by retaining the selected vertices $\vertices_{selected}$ and preserving all edges $\edges_{selected}$ from the original roadmap $\graph$ whose endpoints both belong to the selected vertex set. 
In other words, $\graph_{new}$  corresponds to the induced subgraph of  $\graph$  defined by the selected vertices.


\begin{algorithm}[t]
\caption{Roadmap Reconstruction}  \label{alg:reconstruct}
\begin{algorithmic}[1]
\State \textbf{Input}: Problem instance $I$, trained GNN evaluator $\phi$, threshold $\tau$
\State \textbf{Output} Reconstructed roadmap $\mathcal{G}_{new}$
\Function{ReconstructRoadmap}{$I,\phi,\tau$}
    \State $\mathcal{G} \leftarrow \text{\scshape GenerateDenseRoadmap}(I)$
    \State $\mathcal{G}_{aug} \leftarrow \text{\scshape AugmentGraph}(\mathcal{G})$
    \State $\textbf{y} \leftarrow \phi(\mathcal{G}_{aug})$ \Comment{Vertex importance scores}
    \State $\vertices_{selected} \leftarrow \emptyset$ 
    \For{each vertex $v_i \in \mathcal{G}$}
        \If{$y_i \ge \tau(\graph,\textbf{y})$}
            \State $\vertices_{selected} \leftarrow \vertices_{selected} \cup \{v_i\}$
        \EndIf
    \EndFor
    \State $\edges_{selected} \leftarrow \emptyset$
    \For{each edge $e_{ij} \in \mathcal{G}$}
        \If{$v_i \in V_{selected}  \wedge v_j \in V_{selected}$}
            \State $\edges_{selected}  \leftarrow \edges_{selected}  \cup \{e_{ij}\}$
        \EndIf
    \EndFor
    
    \State $\mathcal{G}_{new} \leftarrow (\vertices_{selected}, \edges_{selected}  )$
    \State return  $\mathcal{G}_{new}$ 
\EndFunction
\end{algorithmic}
\end{algorithm}
\section{Experiments}
In this section, we evaluate our roadmap evaluator and generation in various MAPP problems.

\subsection{Experimental Design}

For all MAPP instances, we model the agent bodies $\radiusspace$ as disk with the same radius $\radius$ in a 2D continuous space ($\classspace = [0,64]^2$) for quick collision detection.
We evaluate with two body-radius setting ($\radius \in \{0, 1\})$) and two kinematic settings: 
\begin{itemize}
    \item continuous-time (ct): each agent can wait or move at the same fixed velocity of 1.
    \item discrete-time (dt): each agent can wait or move to the neighboring vertex in one timestep. 
\end{itemize}
In both kinematic settings, agents move linearly along the roadmap edges.
Under these settings, we generated 25 different MAPP instances with at most 10 randomized start-goal permutations for 1, 2, 4, 8, and 16 agents.
For the MAPF/MAPP algorithm, we used Safety Interval Prioritized Planner (SIPP) \cite{phillips_sipp_2011} as it provides a quick but suboptimal solution and can be used in a discrete or continuous time-space domain.

\textbf{Roadmap Baselines:} To evaluate the effectiveness of our reconstructed roadmaps, we compare across one dense roadmap generation (PRM) as our main results and two sparse roadmap generations (lattice-grid, PRM, and Delaunay Triangulation) as ablation studies.
\begin{itemize}
    \item \textbf{PRM} \cite{kavraki_probabilistic_1996} (Dense): The number of samples was set to 2000 with a minimum and maximum edge length of 0.1 and 5.1, respectively. The maximum number of neighbors per vertex is 15.
    \item \textbf{Lattice Grid} (Sparse): Conventional discretized roadmap used for MAPF with a grid size $[32x32]$
    \item \textbf{Delaunay Triangulation} \cite{kallmann_fully_2004} (Sparse):The number of samples was set to 2000 with a minimum and maximum edge length of 0.1 and 5.1.
\end{itemize}


\subsection{Evaluation Metric}
For the following metrics, we consider the percent increase of the MAPF/MAPP algorithm on our constructed roadmap compared to the original generated roadmap where MAPF/MAPP scenarios were successful on both roadmaps.
\begin{itemize}
    \item \textbf{Sum-of Cost (SOC) Increase}: Solution quality defined in Section \ref{sec:problem_statement} or the average sum of times taken for all agents to reach their tasks
    \item \textbf{Path Length}: The average total travel distance for all agents to reach their tasks
    \item \textbf{Makespan}: The average completion time for the last agent to finish ($\max_{i}T_i$)
    \item \textbf{Runtime}: The average execution time of running the MAPP/MAPF algorithm
    \item \textbf{Number of Nodes Reduction}: The average number of nodes reduced or pruned by the roadmap evaluator.
    \item \textbf{Number of Edges Reduction}: The average number of edges reduced or pruned by the roadmap evaluator.
\end{itemize}

\subsection{Training Setup}
Our dataset comprises 1,800 problem instances on $32\times32$ maps with an obstacle density of 0.1, featuring scenarios with 1, 2, and 4 agents, where we use 1,620 problems for training and 180 for validation.
The number of generated problem permutations per problem instance was at most 64 to create our occupation density map $\heatmap$.
The roadmap generation was done with PRM, sampling 1000 waypoints with a minimum edge length of 0.1, a maximum edge length of 5.1, and a maximum limit of 13 neighbors per node. 
The architectural pipeline begins with two initial linear layers with 64 channels to encode the input, which consists of 4-dimensional node features and 1-dimensional edge features. 
This is followed by three residual Graph Attention Network (GAT) layers, each configured with 4 heads and 16 channels, using Leaky ReLU activation functions. 
A final linear layer with 32 channels reduces the dimensionality to a single scalar output representing the vertex probability score. 
The model is trained using the AdamW \cite{loshchilov_decoupled_2019} optimizer with a learning rate of $0.0001$ and a weight decay of $5\times10^{-4}$. 
We employ a batch size of 128 over 5,000 epochs, incorporating a Laplacian regularization term ($\lambda$) of $0.01$ to ensure spatial smoothness across the roadmap.
We utilize an Intel Ultra 7 265K (20 cores) processor and a single NVIDIA RTX 5090 GPU with 32GB of RAM, where the dataset generation took less than 5 minutes using 20-core multiprocessing, and the training process was completed in approximately 52 minutes.

\subsection{Results}
\begin{figure*}[ht]
    \centering
    \includegraphics[width=0.7\linewidth]{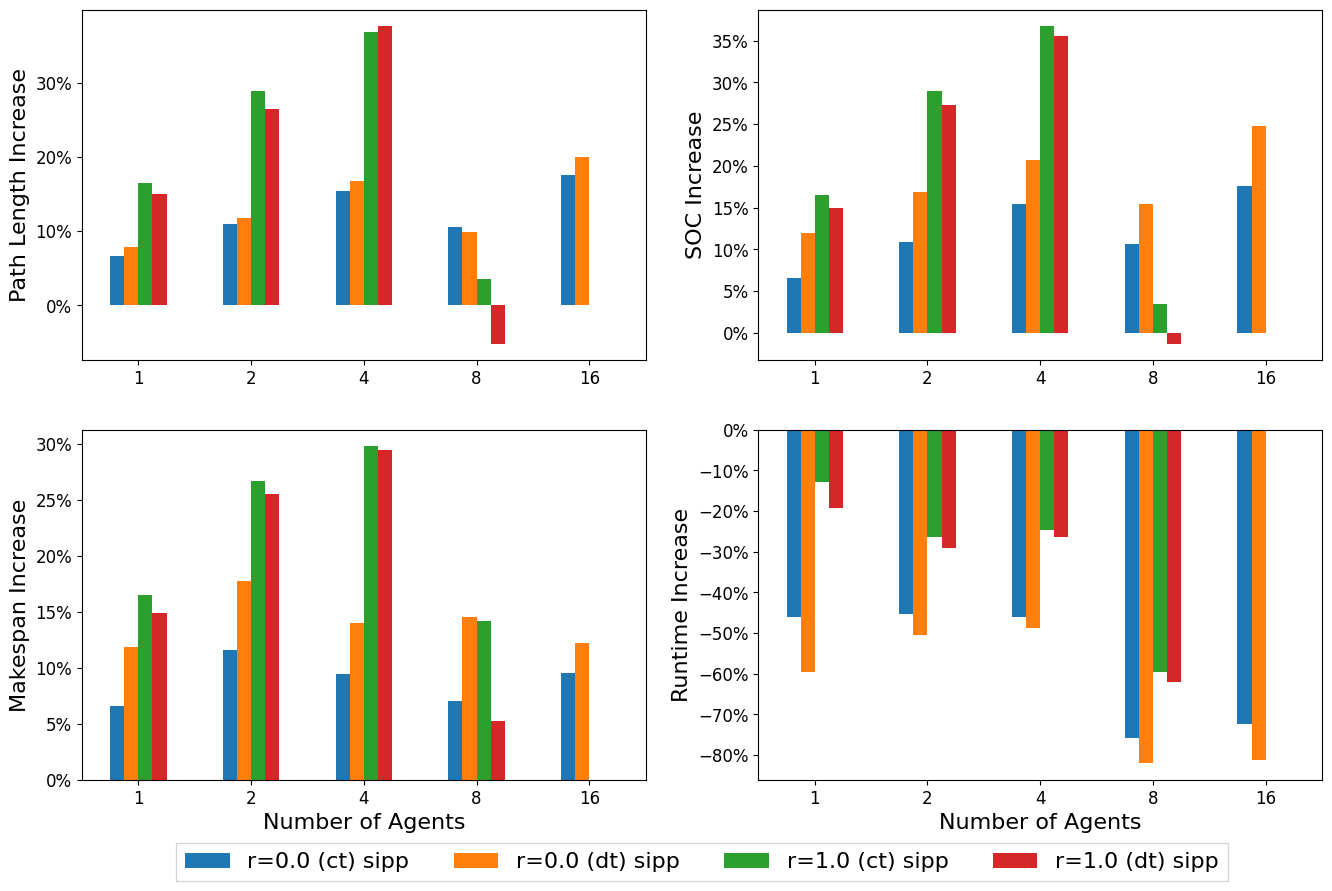}
    \caption{Performance metrics comparing the reconstructed PRM against the original dense PRM across varying agent counts $\numberof = \{1,2,4,8,16\}$.
    Results are shown for points particles ($\radius = 0$, and agents with a defined radius $\radius = 1$ under discrete time ($dt$) and continuous time ($ct$) kinematics.
    While pruning introduces a solution coarseness that increases SOC and path length, it yields a drastic reduction in runtime—up to $80\%$ in high-agent scenarios.
    The 16-agent scenario with $R=1.0$ resulted in a failure to find a solution (N/A in Table I). This suggests that for large-bodied agents in high-density environments, the current thresholding mechanism may prune critical connectivity nodes required for collision avoidance.}
    \label{fig:prm_main}
\end{figure*}

Figure \ref{fig:prm_main} and Table \ref{tab:prm_node_edge_reduction} present our main results. 
Examples of reconstructed roadmaps and their corresponding solutions are shown in Figure \ref{subfig:single_agent_valid}. 
Overall, our roadmap evaluator significantly reduces the search space and planning time for MAPF algorithms.

Solutions found using the reconstructed PRM operate within a smaller search space compared to those generated using the original PRM. 
As shown in Table \ref{tab:prm_node_edge_reduction}, the roadmap evaluator reduces the number of nodes and edges by more than $40\%$ across all agent scenarios and configurations.
Furthermore, planning time is reduced by at least $40\%$ for point-particle agents and by an average of $20\%$ for agents with a radius of 1.

There were instances where the SOC, makespan, and path length were smaller in the sparse reconstructed roadmap than in the original dense roadmap, particularly in the 8-agent PRM scenarios shown in Figure \ref{fig:prm_main} and the 2-agent lattice grid scenarios.
From our observation, in the dense roadmap, some SIPP agents were required to wait for others before reaching their destinations.
In contrast, in the reconstructed roadmap, the agents were able to proceed without waiting.
However, since SIPP does not guarantee global optimality, this behavior may also arise from the non-optimal nature of the prioritized planning process.
Nevertheless, the results suggest that the roadmap evaluator may implicitly identify and remove vertices that contribute to congestion or agent conflicts.

\begin{figure*}[ht]
    \centering
    \begin{subfigure}{.5\textwidth}
        \centering
        \includegraphics[width=\linewidth]{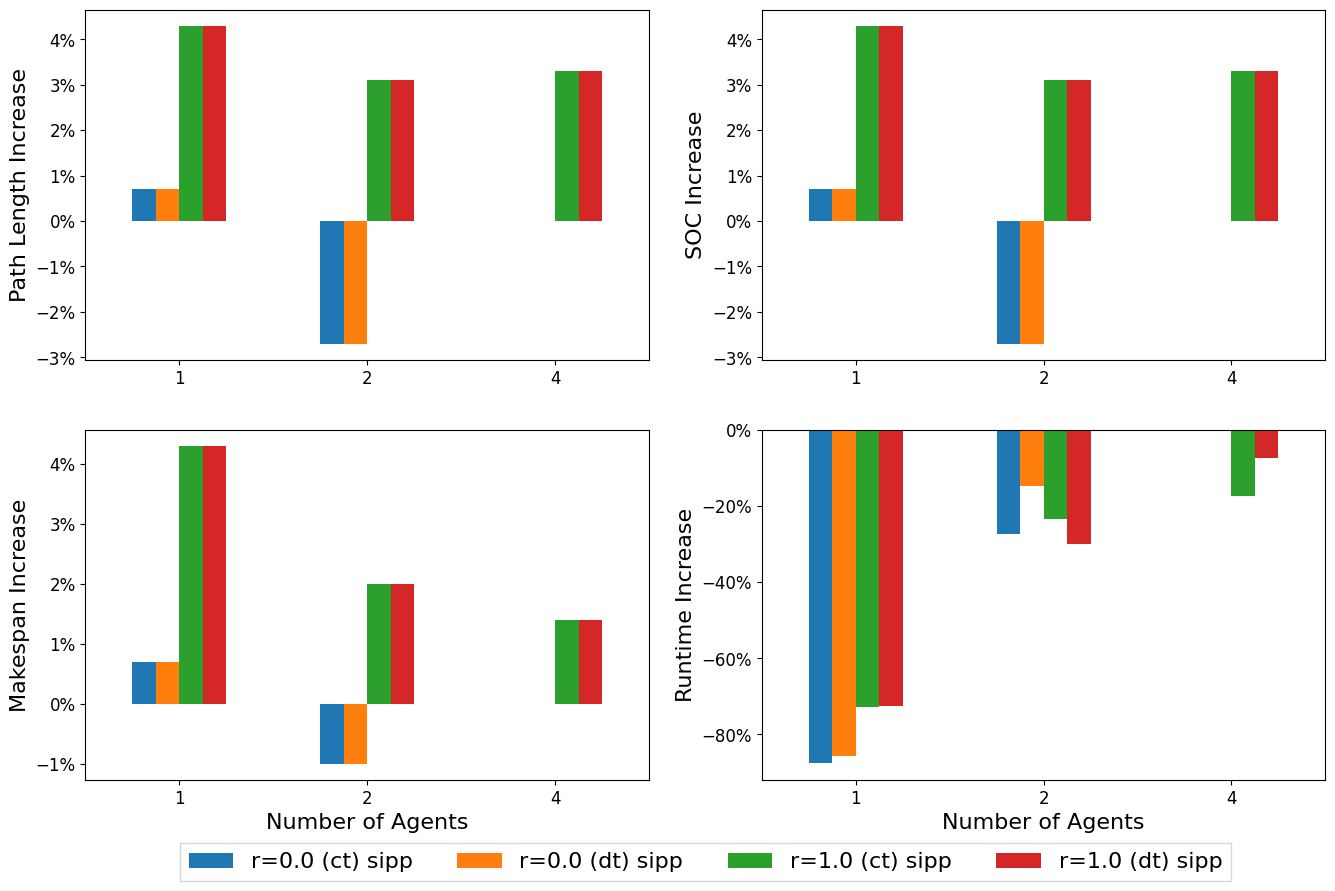}
        \caption{Lattice Grid Ablation}
        \label{subfig:ablation_lattice_grid}
    \end{subfigure}%
    \begin{subfigure}{.5\textwidth}
        \centering
        \includegraphics[width=\linewidth]{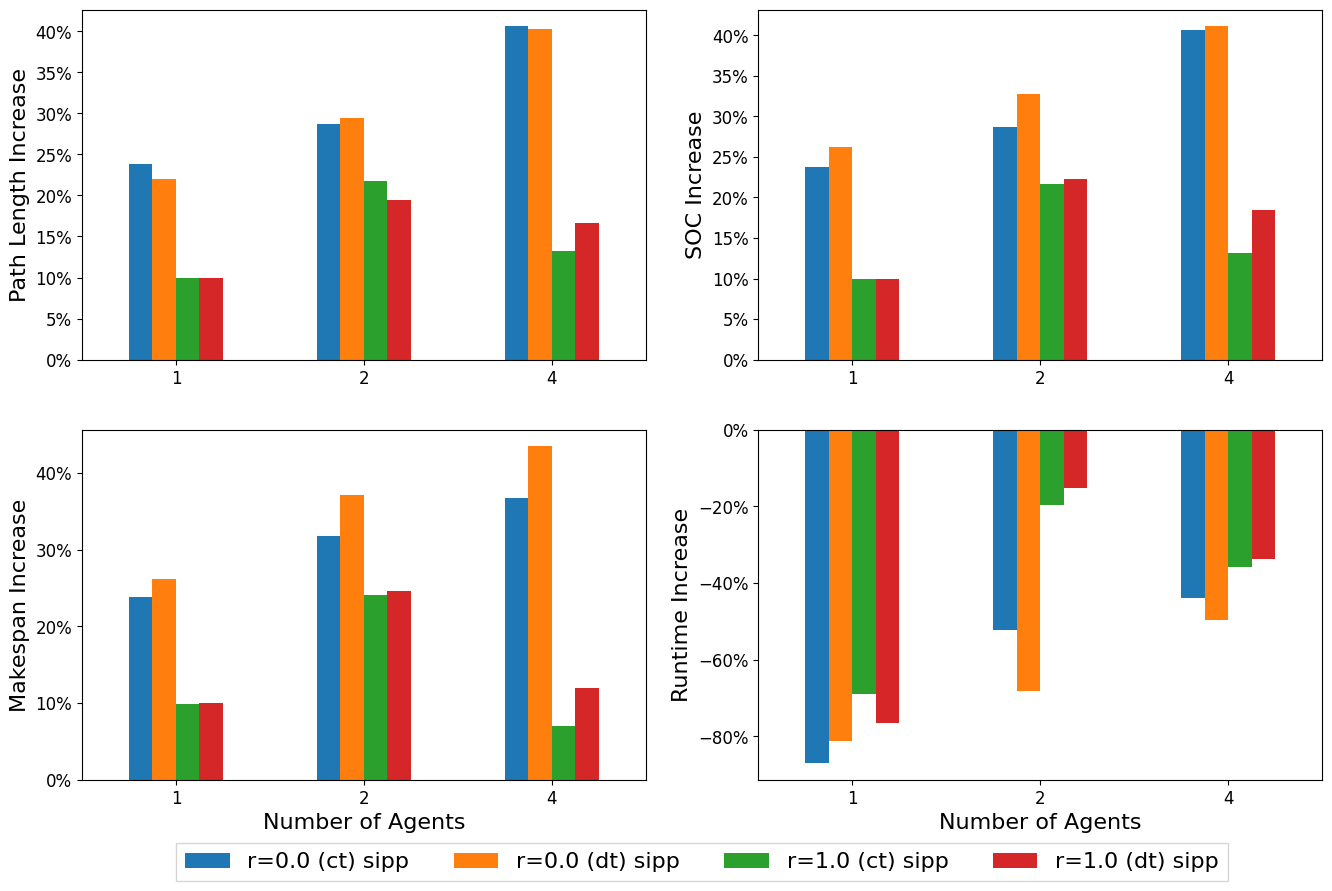}
        \caption{Delaunay Triangulation Ablation}
        \label{subfig:ablation_delaunay_triangulation}
    \end{subfigure}
    \caption{    The charts illustrate the trade-offs when applying the GNN-based roadmap evaluator to inherently sparse graph structures, specifically Lattice Grids (a) and Delaunay Triangulations (b).
    While solutions were found for up to 16 agents on PRMs, these sparse roadmaps were limited to 4-agent scenarios due to the loss of critical connectivity nodes.
    However, both sparse roadmaps experienced significantly shorter runtimes after reconstruction, with reductions reaching up to $80-90\%$ for single-agent cases.
   In the 2-agent Lattice Grid scenario, the reconstructed roadmap yielded lower SOC and path lengths.
    }
    \label{fig:ablation}
\end{figure*}

We further evaluated the capability of the roadmap evaluator on already sparse roadmaps, including lattice grids and Delaunay triangulations, as shown in Figure \ref{fig:ablation}.
Our experiments show that MAPF algorithms operating on the reconstructed roadmaps are able to find solutions for up to four agents. 
While sparse roadmaps can significantly expedite the search process (Table \ref{tab:ablation_sparse}), they also reduce connectivity, which increases the likelihood of failing to find a feasible solution for a large number of agents.

\begin{figure}[ht]
    \centering
    \begin{subfigure}{.23\textwidth}
        \centering
        \includegraphics[width=\linewidth,frame]{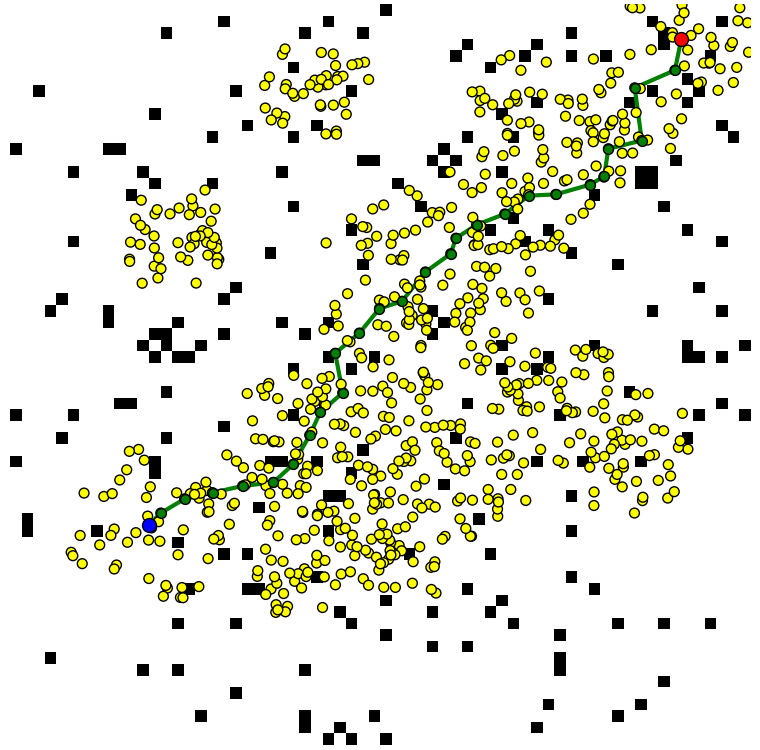}
        \caption{Single Agent Reconstructed PRM Roadmap SIPP Solution}
        \label{subfig:single_agent_prm}
    \end{subfigure}%
    \begin{subfigure}{.23\textwidth}
        \centering
        \includegraphics[width=\linewidth,frame]{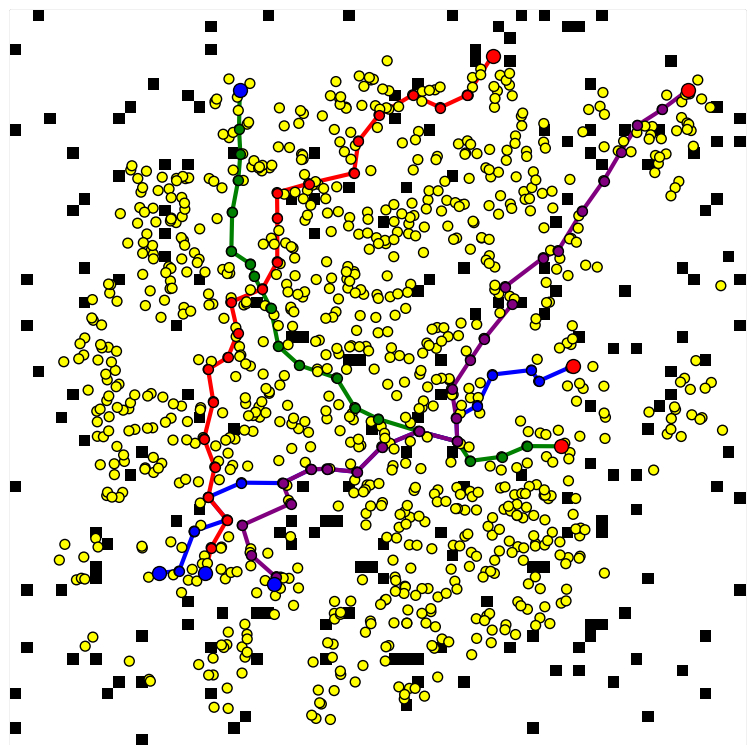}
        \caption{Multi-Agent Reconstructed PRM Roadmap SIPP Solution}
        \label{subfig:multi_agent_prm}
    \end{subfigure}
    \caption{The colorful lines are the path in the solution for each agent in the PRM reconstructed map in a 64x64 square map.}
    \label{fig:prm_rest}
\end{figure}

\begin{table}[]
\centering
\begin{tabular}{|c|c|rr|rr|}
\hline
\multirow{2}{*}{\begin{tabular}[c]{@{}c@{}}Dense\\ Roadmap\end{tabular}} & \multirow{2}{*}{\begin{tabular}[c]{@{}c@{}}Number of \\ Agents\end{tabular}} & \multicolumn{2}{c|}{\begin{tabular}[c]{@{}c@{}}Number of Nodes\\ Reduction\end{tabular}} & \multicolumn{2}{c|}{\begin{tabular}[c]{@{}c@{}}Number of Edges\\ Reduction\end{tabular}} \\ \cline{3-6} 
                                                                         &                                                                              & \multicolumn{1}{c|}{R = 0.0}                 & \multicolumn{1}{c|}{R=1.0}                & \multicolumn{1}{c|}{R = 0.0}                 & \multicolumn{1}{c|}{R=1.0}                \\ \hline
\multirow{5}{*}{PRM}                                                     & 1                                                                            & \multicolumn{1}{r|}{40.7\%}                  & 38.0\%                                    & \multicolumn{1}{r|}{46.5\%}                  & 42.0\%                                    \\ \cline{2-6} 
                                                                         & 2                                                                            & \multicolumn{1}{r|}{41.5\%}                  & 39.0\%                                    & \multicolumn{1}{r|}{48.0\%}                  & 43.8\%                                    \\ \cline{2-6} 
                                                                         & 4                                                                            & \multicolumn{1}{r|}{42.5\%}                  & 39.6\%                                    & \multicolumn{1}{r|}{49.5\%}                  & 44.3\%                                    \\ \cline{2-6} 
                                                                         & 8                                                                            & \multicolumn{1}{r|}{44.2\%}                  & 40.8\%                                    & \multicolumn{1}{r|}{51.9\%}                  & 44.8\%                                    \\ \cline{2-6} 
                                                                         & 16                                                                           & \multicolumn{1}{r|}{43.2\%}                  & \multicolumn{1}{c|}{N/A}                  & \multicolumn{1}{r|}{50.9\%}                  & \multicolumn{1}{c|}{N/A}                  \\ \hline
\end{tabular}
\caption{The number of nodes and edges reduction reaches at least $40\%$ except for $R=1.0$ for 16 agents due to no solution founded under the graph.}
\label{tab:prm_node_edge_reduction}
\end{table}

\begin{table}[]
\begin{tabular}{|c|c|cc|cc|}
\hline
\multirow{2}{*}{\begin{tabular}[c]{@{}c@{}}Sparse\\ Roadmap\end{tabular}}         & \multirow{2}{*}{\begin{tabular}[c]{@{}c@{}}Number of\\ Agents\end{tabular}} & \multicolumn{2}{c|}{\begin{tabular}[c]{@{}c@{}}Number of Nodes\\ Reduction\end{tabular}} & \multicolumn{2}{c|}{\begin{tabular}[c]{@{}c@{}}Number of Edges\\ Reduction\end{tabular}} \\ \cline{3-6} 
                                                                                  &                                                                             & \multicolumn{1}{c|}{R = 0.0}                           & R=1.0                           & \multicolumn{1}{c|}{R=0.0}                            & R=1.0                            \\ \hline
\multirow{3}{*}{\begin{tabular}[c]{@{}c@{}}Lattice\\ Grid\end{tabular}}           & 1                                                                           & \multicolumn{1}{c|}{33.7\%}                            & 33.9\%                          & \multicolumn{1}{c|}{34.5\%}                           & 34.7\%                           \\ \cline{2-6} 
                                                                                  & 2                                                                           & \multicolumn{1}{c|}{34.0\%}                            & 33.7\%                          & \multicolumn{1}{c|}{34.9\%}                           & 34.5\%                           \\ \cline{2-6} 
                                                                                  & 4                                                                           & \multicolumn{1}{c|}{33.1\%}                            & N/A                             & \multicolumn{1}{c|}{34.1\%}                           & N/A                              \\ \hline
\multirow{3}{*}{\begin{tabular}[c]{@{}c@{}}Delaunay\\ Triangulation\end{tabular}} & 1                                                                           & \multicolumn{1}{c|}{38.5\%}                            & 36.2\%                          & \multicolumn{1}{c|}{44.8\%}                           & 40.3\%                           \\ \cline{2-6} 
                                                                                  & 2                                                                           & \multicolumn{1}{c|}{38.8\%}                            & 37.3\%                          & \multicolumn{1}{c|}{45.5\%}                           & 42.2\%                           \\ \cline{2-6} 
                                                                                  & 4                                                                           & \multicolumn{1}{c|}{38.8\%}                            & 36.8\%                          & \multicolumn{1}{c|}{46.1\%}                           & 41.7\%                           \\ \hline
\end{tabular}
\caption{The number of nodes and edges reduction reaches at least $30\%$ for upto 4 agents as pruning on sparse graph can lead to no solutions with a large number of agents and indicated by N/A.}
\label{tab:ablation_sparse}
\end{table}

\subsection{Limitation}
\vspace{-5pt}
Although the proposed GNN-based roadmap evaluator is capable of identifying key regions of interest and pruning redundant waypoints to produce a sparser roadmap, it has a few limitations.
First, if the initial roadmap is not sufficiently dense, the evaluator may prune vertices that are essential for connectivity, thereby eliminating feasible paths between start–goal pairs.
Second, the reconstruction process relies on a thresholding function applied to the GNN output scores. This threshold acts as a tunable hyperparameter that directly controls the sparsity of the reconstructed roadmap.
A higher threshold produces a sparser graph, reducing the search space but increasing the risk of removing critical vertices and destroying connectivity.
Conversely, a lower threshold preserves connectivity more reliably but retains a larger search space, limiting computational gains.
In practice, the effectiveness of the proposed method depends on selecting an appropriate threshold value that balances sparsity and connectivity.
\section{Conclusion}
We presented a heterogeneous Graph Neural Network (GNN) framework for roadmap generation and evaluation, introducing a novel graph representation tailored for shared-roadmap construction in multi-agent path planning. 
The proposed GNN-based roadmap evaluator assigns importance scores to vertices in an initially dense roadmap and prunes those deemed unnecessary via a thresholding function.
Through its training procedure—leveraging multiple permutations of the same problem instance—and by encoding start and goal configurations as landmarks, the GNN captures structural patterns that generalize across agent assignments. 
As a result, the reconstructed roadmap is task-agnostic, enabling MAPF algorithm to reuse the same map under different task assignments, making it suitable for multi-agent pick-up and delivery (MAPD) tasks.
We demonstrated that our roadmap evaluator reduced the search effort of MAPF/MAPP algorithms, reaching at least $40\%$ reduction in runtime and graph size for dense roadmaps, and potential remove points of congestion or conflict.
In future work, we plan to replace the manually tuned thresholding mechanism with a learned thresholding function and extend the framework to explicitly address MAPD problems.
Such an extension would enable adaptive, instance-aware pruning, reducing the likelihood of disconnecting critical regions of the roadmap and thereby minimizing the risk of infeasible solutions.

\vspace{-5pt}
\bibliographystyle{IEEEtran}
{
\small
\bibliography{references}
}
\end{document}